\documentclass[runningheads]{llncs}

\usepackage{eccv}

\usepackage{eccvabbrv}

\usepackage{graphicx}
\usepackage{booktabs}
\usepackage{array}        
\usepackage{subcaption}   
\usepackage{xcolor}  
\usepackage{enumitem}
\usepackage{tabularx}
\usepackage{makecell}

\usepackage[accsupp]{axessibility}  

\usepackage{hyperref}

\usepackage{orcidlink}

\begin{document}

\title{SVGEval: A Vision-Grounded Framework for Perceptual-Quality Benchmarking and Evaluation in Text-to-SVG Generation} 

\titlerunning{SVGEval}

\author{Yiming Wang\inst{2,3}\textsuperscript{*} \and
Ye Chen\inst{1,2}\textsuperscript{*} \and
Hanqi Chen\inst{2} \and
Bingbing Ni\inst{1,2}\textsuperscript{$\dagger$}
}

\authorrunning{Y. Wang et al.}

\institute{
Guangdong Laboratory of Artificial Intelligence and Digital Economy (SZ), Shenzhen 518107, China\\
\and
School of Integrated Circuits (School of Information Science and Electronic Engineering), Shanghai Jiao Tong University, Shanghai, China\\
\and
Zhiyuan College, Shanghai Jiao Tong University, Shanghai, China\\
\email{\{javadcc1, chenye123, caffinities, nibingbing\}@sjtu.edu.cn}
}

\newcolumntype{M}[1]{>{\centering\arraybackslash}m{#1}}
\newcolumntype{L}[1]{>{\raggedright\arraybackslash}m{#1}}
\newcolumntype{C}[1]{>{\centering\arraybackslash}p{#1}}
\newcolumntype{P}[1]{>{\raggedright\arraybackslash}p{#1}}
\newcommand{\svgthumb}[1]{%
  \centering
  \includegraphics[height=1.45cm,keepaspectratio]{#1}%
}
\newcommand{\myrunin}[1]{\par\smallskip\noindent\textbf{#1.}\ }

\maketitle

\begingroup
\renewcommand{\thefootnote}{%
  \ifcase\value{footnote}%
  \or *%
  \or \textdagger%
  \fi}
\footnotetext[1]{Equal contribution.}
\footnotetext[2]{Corresponding author: Bingbing Ni.}
\endgroup

\begin{abstract}

Multimodal large models are increasingly used to generate scalable vector graphics (SVG), but reliable evaluation remains underexplored. Existing protocols are often code-centric or borrow raster-image metrics after rendering SVGs, which fail to reflect human perception and overlook SVG-specific qualities such as geometry and spatial composition. We introduce SVGEval, a vision-grounded multimodal benchmark for human-aligned SVG quality assessment. SVGEval explicitly incorporates visual renderings to evaluate whether models can judge the rendered outcome rather than only inspect SVG code, and provides high-quality annotations obtained via multi-round human labeling with expert refinement. Systematic evaluations across representative multimodal models reveal a clear gap: models perform relatively well on semantic alignment and aesthetics, yet struggle on geometry- and layout-related judgments. Building on SVGEval, we train an explainable SVG quality scorer that outputs multi-aspect scores with textual rationales. Ablations show that explicit visual grounding and reasoning supervision are crucial, especially for spatial and geometric assessment. SVGEval offers a reliable testbed and practical scorer for evaluating and improving SVG generation in the era of multimodal models.

\end{abstract}

\section{Introduction}
\label{sec:intro}
Scalable Vector Graphics (SVG) has become a fundamental visual format for modern digital content due to its resolution-independence, compact storage, and high editability \cite{svg2}. Unlike raster images that represent appearance as dense pixels, SVG encodes visual content as a structured program consisting of paths, primitives, and hierarchical groups. This representation makes SVG especially attractive for applications where geometric fidelity and reusability matter, such as icons, diagrams, UI design, and infographic authoring.

Driven by the recent progress of multimodal large models (MLLMs), text-to-SVG generation has advanced rapidly \cite{starvector,omnisvg,llm4svg,Chen_2025}, enabling models to produce complex vector graphics from natural language instructions. While the generation capability is improving at an impressive pace, evaluation lags behind. In contrast to text-to-image generation, where widely adopted quantitative metrics and protocols have been extensively studied, SVG generation still lacks a unified and reliable evaluation standard. As a result, comparisons across methods are often inconsistent, and improvements reported under one evaluation setting may not translate to human-perceived quality in practice.

A common evaluation practice today is to treat SVG as a raster image by rendering it into a bitmap and then applying image-generation metrics such as MSE, LPIPS \cite{lpips}, FID \cite{fid}, and CLIPScore \cite{clipscore, clip} to quantify fidelity and prompt alignment. However, this raster-centric paradigm is fundamentally misaligned with the nature of SVG. SVG quality is largely determined by structured geometry and spatial composition rather than local textures, and many perceptual differences that are critical for vector graphics—e.g., slight geometric distortions, misaligned spacing, broken symmetries, or inconsistent path structures—may be weakly reflected (or even ignored) by pixel-level errors and deep feature similarities. Conversely, rasterization introduces arbitrary factors (resolution, anti-aliasing, stroke rendering, and sampling) that can dominate MSE/LPIPS/FID without corresponding to meaningful changes in the underlying vector structure. Similarly, CLIP-based similarity mainly captures coarse semantic alignment at the image level, but is often insensitive to fine-grained layout rationality and geometric correctness that strongly affect the usability and perceived quality of SVG \cite{spatialclip, aro}. 

These limitations suggest that evaluating SVG generation requires criteria and protocols that explicitly consider SVG’s vector-specific structured attributes. In this work, we propose SVGEval, a vision-grounded framework for benchmarking and evaluating the perceptual quality of text-to-SVG generation with multimodal models. Our key premise is that SVG is ultimately consumed visually by humans; therefore, an effective evaluator should be able to judge the rendered visual outcome while also accounting for SVG-specific structural properties.

Concretely, based on an analysis of SVG characteristics, we introduce an evaluation protocol that covers four complementary aspects: semantic consistency with the text instruction, geometric structure quality (e.g., shape integrity and structural correctness), spatial layout rationality (e.g., alignment, spacing, and composition), and visual aesthetics. Building on this protocol, we construct a two-stage, difficulty-progressive mixed-task benchmark. The first stage focuses on coarse-grained verification: models perform binary (yes/no) judgments to test basic perception and understanding of critical SVG attributes. The second stage requires fine-grained assessment: models provide quantitative ratings for each aspect, directly probing nuanced evaluative ability beyond coarse correctness. Importantly, SVGEval is explicitly vision-grounded: each example includes the SVG representation together with its visual rendering, so models are evaluated on what users actually see rather than purely on code semantics. All annotations are collected through multi-round human labeling and further expert refinement, resulting in high-quality judgments that strongly align with human preferences and enable reliable benchmarking across models.

Using SVGEval, we conduct systematic evaluations of representative state-of-the-art multimodal large models. Our results reveal a consistent and practically important gap: models generally perform better on semantic consistency and overall aesthetics, yet struggle significantly on spatial reasoning and geometric structure quality. This indicates that current models remain limited in assessing SVG-specific structured attributes, and also explains why raster-centric metrics can be misleading for vector graphics: they often fail to surface precisely the failure modes that matter most for SVG usability and perceived quality.

Beyond benchmarking, we further leverage SVGEval to train an explainable SVG quality scorer. Compared with counterparts that formulate SVG evaluation as a source-code understanding or code-centric task \cite{vgbench,vcode,svgenius,starvector}, our scorer incorporates explicit visual renderings and employs chain-of-thought style reasoning supervision to better align the evaluation process with human judgment. The scorer outputs calibrated quantitative scores for each aspect together with textual rationales, providing not only a final rating but also interpretable feedback. Extensive experiments and ablations demonstrate that both explicit SVG modality information and reasoning supervision are crucial for improving evaluation performance—especially on the geometry- and layout-related aspects—highlighting the necessity of multimodal, vision-centric evaluation for vector graphics.

We hope that SVGEval and the proposed scorer will provide a reliable reference for benchmarking and evaluating text-to-SVG generation in the era of multimodal large models, and inspire future research on structure-aware SVG generation and human-aligned evaluation.

Our contributions can be summarized as:
(1) We introduce SVGEval, a vision-grounded framework that supports benchmarking and evaluating the perceptual quality of text-to-SVG generation with multimodal models.
(2) We design a human-aligned evaluation protocol covering semantic and SVG-specific structured aspects, and construct a difficulty-progressive benchmark with high-quality multi-round human annotations and expert refinement.
(3) We provide a systematic study across major multimodal models and reveal a consistent weakness in geometry and spatial-layout assessment.
(4) We train an explainable SVG quality scorer with visual grounding and reasoning supervision, and show through ablations that explicit SVG modality inputs and chain-of-thought supervision are key to accurate evaluation.

\section{Related Works}

\myrunin{Vector Graphics Generation}
Recent progress in large language models and multimodal LLMs has enabled direct generation in the \emph{SVG code space}.
StarVector introduces a multimodal model for SVG generation and proposes SVG-Bench for task-oriented evaluation \cite{starvector}.
OmniSVG further explores a unified framework for multimodal SVG generation with large-scale training data \cite{omnisvg}.
LLM4SVG improves LLMs' understanding and generation of complex vector graphics via structured representations \cite{llm4svg}.
SVGThinker emphasizes instruction alignment and reasoning-driven text-to-SVG generation to improve robustness and controllability \cite{Chen_2025}.
While these methods advance SVG synthesis, their evaluations are often task-completion oriented or rely on generic raster-image metrics after rendering, which provide limited diagnosis for SVG-specific geometric and layout defects.

\myrunin{Benchmarks for Visual Perception and Quality Understanding}
A broad line of research develops general-purpose benchmarks to systematically evaluate MLLMs' vision-language capabilities, spanning \emph{visual perception} and higher-level \emph{cognition/reasoning}.
For example, MME measures both perception and cognition abilities across a set of diverse subtasks \cite{mme}.
Within this broader evaluation landscape, Q-Bench focuses on low-level visual perception and image-quality understanding\cite{qbench}.
Q-Bench+ further extends this setting from single images to image pairs, enabling comparison-based assessment that better matches human quality judgments \cite{qbenchplus}.
Our work aligns with these efforts in probing perceptual judgment, but targets \emph{SVG renderings} and explicitly models SVG-critical dimensions such as \emph{spatial layout} and \emph{structural quality}.

\myrunin{Benchmarks for SVG/Vector Understanding, Editing, and Generation}
Several benchmarks evaluate LLMs/VLMs on vector-graphics processing.
VGBench assesses LLMs on vector graphics understanding and generation across formats \cite{vgbench}.
SVGenius benchmarks SVG processing with tasks spanning understanding, editing, and generation \cite{svgenius}.
VCode reframes multimodal understanding as generating SVG code that preserves symbolic meaning \cite{vcode}.
VectorGym provides a multi-task benchmark suite for SVG generation and manipulation aligned with real-world workflows \cite{vectorgym}.
Compared with these largely task-driven settings, SVGEval focuses on \emph{quality diagnosis}: it defines a multi-aspect rubric over rendered outcomes and supports training an explainable scorer for stable, evidence-grounded assessment.

\myrunin{LLM-as-a-Judge and Rubric-Guided Evaluators}
LLM-based judges have been widely used to approximate human preferences and scale evaluation.
MT-Bench/Chatbot Arena studies the effectiveness and biases of LLM-as-a-judge \cite{mtbench}.
G-Eval proposes rubric-based evaluation with structured scoring procedures \cite{geval}.
Prometheus trains an open evaluator model that performs fine-grained rubric-guided assessment \cite{prometheus}.
We adopt the rubric-guided evaluation philosophy but specialize it to SVG quality assessment: our scorer takes (prompt, rendering, SVG code) as input and outputs dimension-wise scores with evidence-grounded rationales, targeting SVG-specific spatial and structural defects.

\begin{figure*}[t]
  \centering
  \includegraphics[width=0.94\textwidth]{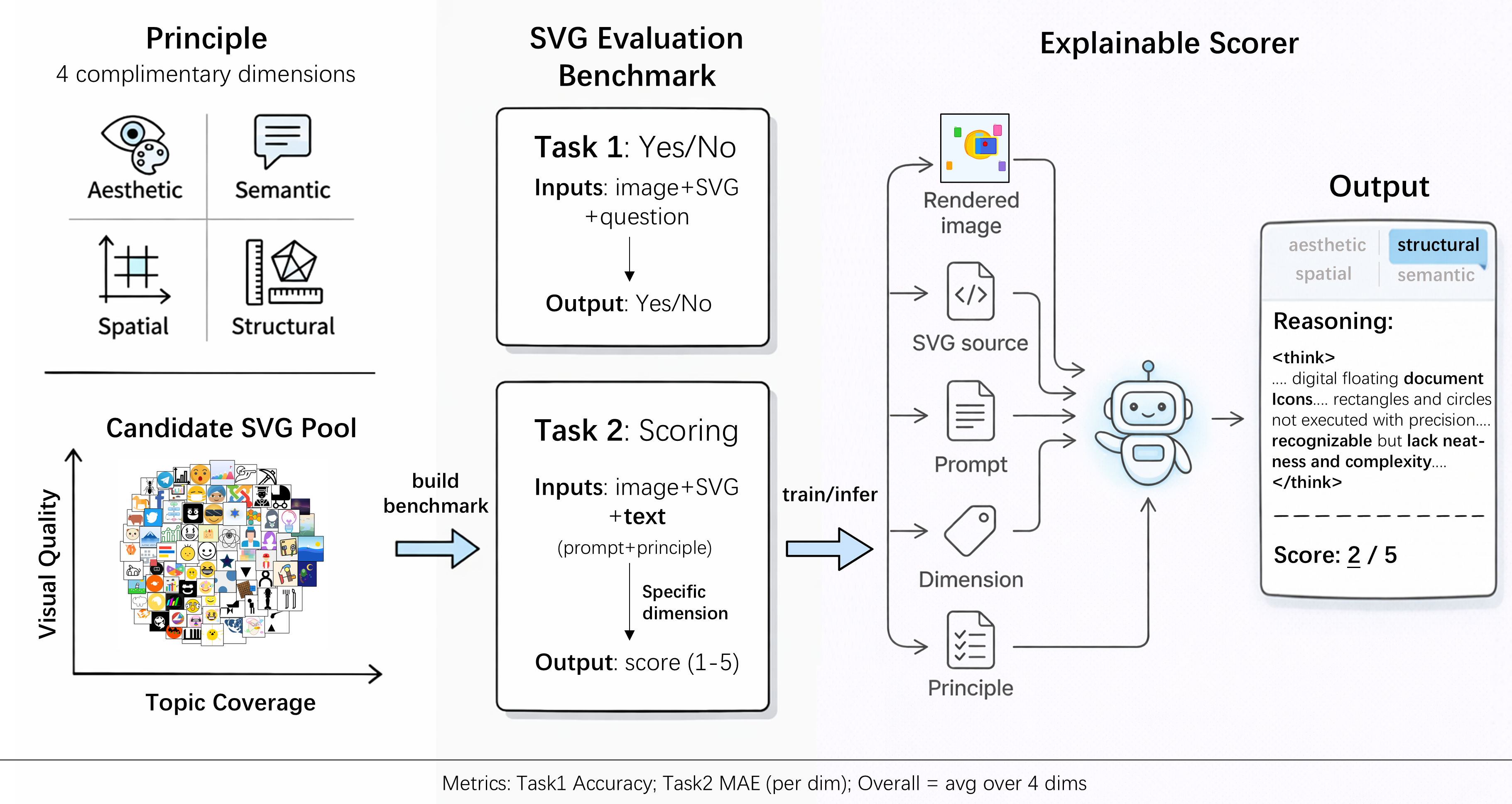}
  \caption{\textbf{Overview of SVGEval.} We define a four-aspect rubric and build a two-part benchmark (binary diagnosis and 1--5 scoring) with visual grounding. SVGEval further enables training an explainable scorer that predicts aspect-wise scores with evidence-grounded rationales from the prompt, rendered image, and SVG code.}
  \label{fig:pipeline}
\end{figure*}

\section{SVGEval Benchmark}

\subsection{Evaluation Principle}
\label{sec:principle}

\myrunin{Evaluation Target} Although SVG is represented as a program, the perceived generation quality is ultimately determined by the rendered outcome. SVGEval therefore treats the \textbf{rasterized rendering} as the primary evaluation target: all criteria are defined over what humans perceive in the final image, rather than over SVG syntax validity or code elegance.

At the same time, SVGEval is \textbf{not} identical to standard text-to-image evaluation. In real workflows, models can often access both the rendered image and the underlying structured representation. To reflect this setting and enable more reliable diagnosis, we allow the evaluator model to see \textbf{(i) the prompt, (ii) the rendered image, and (iii) the SVG code} during judgment, while the ground-truth labels are defined with respect to the visual rendering.

\myrunin{Four Complementary Dimensions} We design four complementary dimensions that aim to capture distinct aspects of rendered quality and support fine-grained diagnostic analysis.
\begin{itemize}[leftmargin=*, topsep=2pt, itemsep=2pt, parsep=0pt]
    \item \textbf{Aesthetic}: visual clarity and overall presentation quality, including color harmony, composition, style coherence, and perceptual refinement. This dimension targets “how good it looks” beyond strict correctness.
    \item \textbf{Semantic}: conceptual alignment with the text prompt only, independent of how well it is drawn.
    \item \textbf{Spatial}: 2D layout correctness on the canvas—the accuracy of each element’s absolute position, the accuracy of relative positions among elements, and the overall spatial harmony of the global layout.
    \item \textbf{Structural}: geometric integrity and construction precision—closed paths, continuity, shape completeness, engineering-level correctness of geometry, and the balance between complexity and neatness.
\end{itemize}
These dimensions are designed to capture distinct perceptual aspects of rendered quality. For example, a geometrically perfect shape can still be semantically incorrect (high Structural, low Semantic), and two individually correct objects can be spatially misplaced (high Structural, low Spatial). This separation facilitates more interpretable attribution of failures to different perceptual aspects, rather than collapsing everything into a single “quality” score.

\myrunin{Hierarchical Attribution Strategy}
Although the four dimensions are defined to reflect different evaluation focuses, perceptual judgments may naturally exhibit overlap across aspects. In particular, Aesthetic impressions often co-occur with geometry- and layout-related defects (e.g., broken shapes and chaotic layouts tend to degrade overall visual appeal). To reduce ambiguity in multi-aspect scoring, SVGEval adopts a hierarchical attribution principle:

{\setlength{\topsep}{2pt}
 \setlength{\itemsep}{2pt}
 \setlength{\parsep}{0pt}
 \begin{itemize}
   \item \textbf{Explicit errors first.} If a negative visual impression is clearly caused by geometric defects (e.g., gaps, broken strokes, unclosed paths) or spatial defects (e.g., unintended overlaps, off-canvas placement), the penalty is attributed to \textbf{Structural} or \textbf{Spatial} first.
   \item \textbf{Residual definition of Aesthetic.} Aesthetic is reserved for visual properties not well captured by Structural/Spatial correctness, such as style maturity, pleasing palette, smoothness/cleanliness, and overall “finish”.
 \end{itemize}
}

This strategy prevents Aesthetic from becoming a catch-all bucket and improves the interpretability of multi-aspect evaluation.

\subsection{Benchmark Construction}

\begin{figure*}[t]
  \centering
  \begin{subfigure}{0.47\textwidth}
    \centering
    \includegraphics[width=0.97\linewidth]{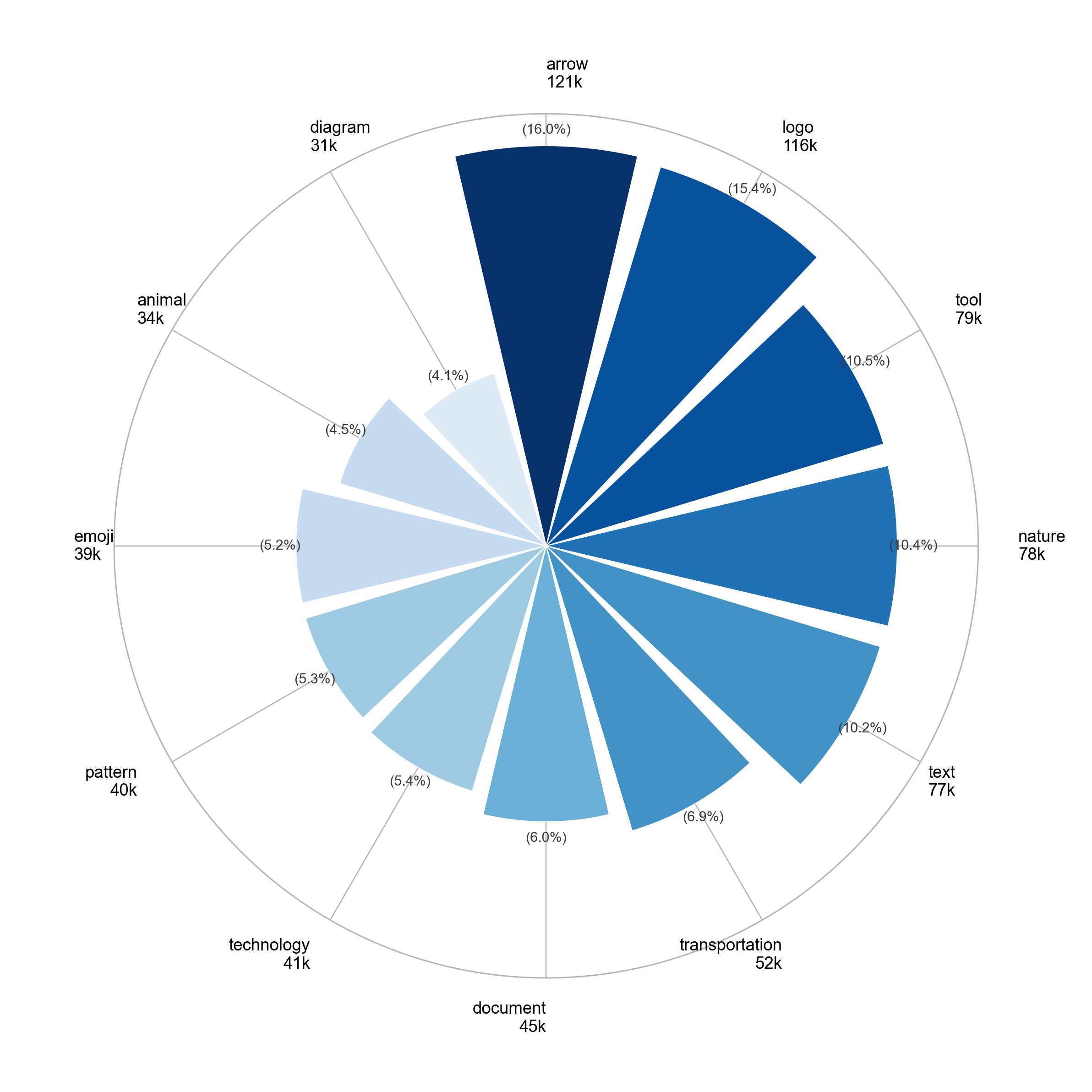}
    \caption{\textbf{Topic distribution} of the 1,000 sampled prompts (used for stratified sampling).}
    \label{fig:topic}
  \end{subfigure}\hfill
  \begin{subfigure}{0.47\textwidth}
    \centering
    \includegraphics[width=0.97\linewidth]{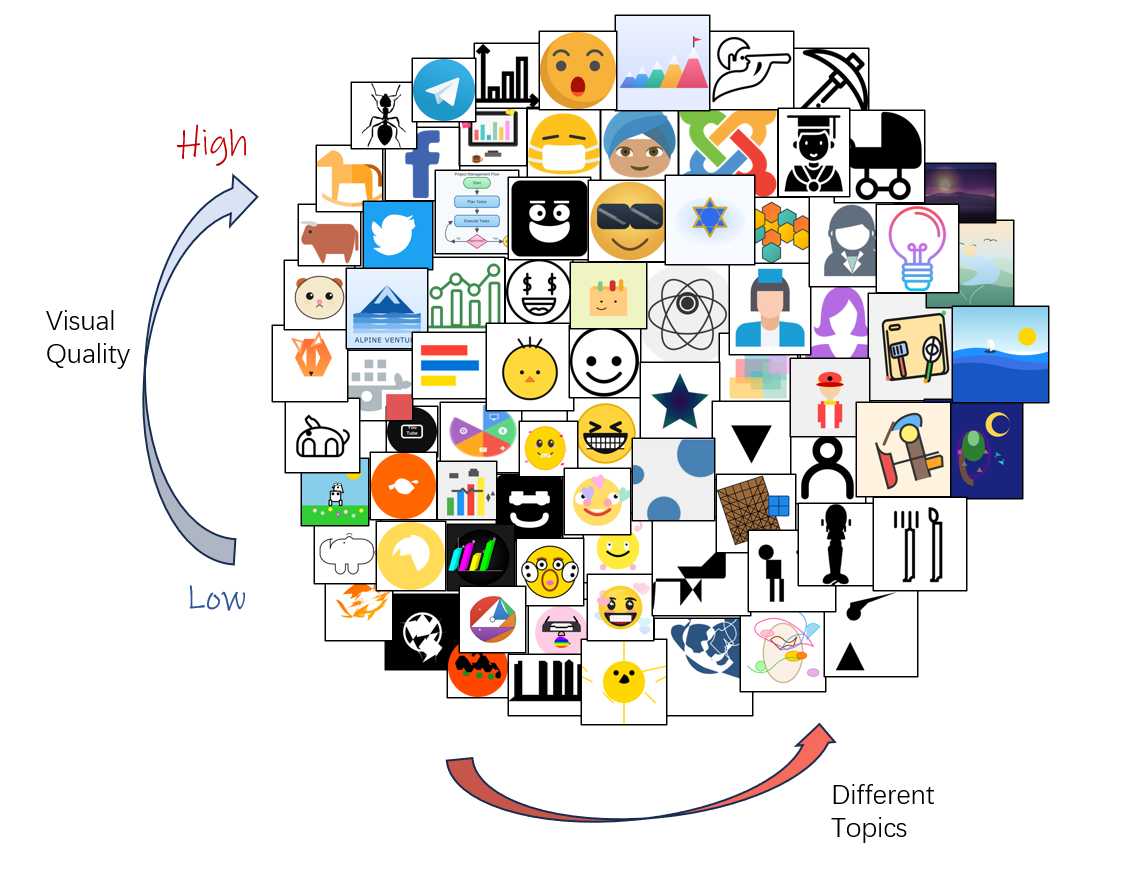}
    \caption{\textbf{Candidate pool} spanning diverse topics and a wide quality spectrum, mixing real SVGs and model generations.}
    \label{fig:candidate_pool}
  \end{subfigure}
  \caption{\textbf{Data coverage of SVGEval.} Left: topic-aware prompt sampling improves category coverage. Right: the candidate pool provides a continuous quality gradient for benchmarking and scorer training.}
  \label{fig:data_overview}
\end{figure*}

\myrunin{Data Engine}
To approximate real-world SVG usage, we construct a large-scale corpus of real SVG--text pairs (sourced from StarVector \cite{starvector} and SVG Repo \cite{svgrepo}). 
Since existing SVG benchmarks often suffer from category bias, we adopt a \textbf{topic-aware prompt sampling} strategy.
Specifically, we use an LLM to extract topic descriptors and cluster the corpus into semantic groups, obtaining an empirical topic distribution, as shown in \cref{fig:topic}.
We then perform stratified sampling following this distribution to collect \textbf{1,000 prompts}, covering diverse scenarios ranging from simple icons to complex diagrams.

\myrunin{Candidate Pool}
To cover a full spectrum of SVG quality, we build a candidate pool of roughly \textbf{10k} samples by mixing (i) generations from several representative LLMs/MLLMs with varying abilities and (ii) real SVGs from the corpus.
The mixture serves two purposes: (1) model-generated SVGs provide diverse failure patterns (especially spatial and structural defects), forming a continuous quality gradient from severely broken to near-correct; 
(2) real SVGs act as \textbf{high-quality anchors}, providing a practical upper bound that current models rarely reach.

\myrunin{Two-Tier Design}
Based on the candidate pool, SVGEval is organized into two complementary parts (each containing \textbf{500} instances) with different output formats and curation protocols. 
A key design goal is to (i) mitigate model-specific response bias through balanced supervision signals, and (ii) ensure that the benchmark supports fine-grained diagnosis via high-quality human annotations.

{\setlength{\topsep}{2pt}
 \setlength{\itemsep}{2pt}
 \setlength{\parsep}{0pt}
 \begin{itemize}
   \item \textbf{Part I: Perception / Yes--No.}
   Following the evaluation principle discussed in \cref{sec:principle}, We construct a set of binary diagnostic questions that probe basic perceptual properties.
   Concretely, we first use a multimodal model to propose candidate questions and preliminary answers given the prompt, rendering, and SVG code, and then \textbf{human annotators revise and validate} both the questions and the Yes/No labels to ensure correctness and clarity.
   To avoid the potential bias in LLM that may skew towards either \textit{yes} or \textit{no} response, we explicitly curate the dataset to maintain an approximately \textbf{balanced positive/negative ratio} (close to \textbf{1:1}) across question types.

    \item \textbf{Part II: Quality Assessment / Scoring.}
    This part quantifies SVG quality on the four dimensions using a \textbf{1--5 Likert scale}.
    To make the benchmark discriminative across different quality regimes, we adjust the sampling ratio(concretely we adopt 1:4) between real SVGs and model-generated SVGs when selecting from the candidate pool, such that the resulting benchmark exhibits a \textbf{relatively balanced score distribution} across score bins for each dimension.
    Each instance is \textbf{independently scored by multiple annotators} on all dimensions, and disagreements are resolved through \textbf{expert adjudication} to produce the final gold scores, improving both consistency and reliability.
 \end{itemize}
}
When benchmarking, the model can access the rendered images, SVG source code, and related prompts simultaneously.

\subsection{Evaluation Metrics}

\myrunin{Part I: Perception Metrics}
Part I is a binary classification task. 
We report \textbf{Accuracy} as the primary metric, measuring the proportion of correct Yes/No judgments (overall and optionally per question type).

\myrunin{Part II: Scoring Metrics}
Part II predicts ordinal scores $S_{\text{pred}} \in \{1,\dots,5\}$ for each dimension against human annotations $S_{\text{gt}}$.
Since the Likert scale is ordinal and minor disagreements are expected in subjective ratings, we report two complementary metrics:

{\setlength{\topsep}{2pt}
 \setlength{\itemsep}{2pt}
 \setlength{\parsep}{0pt}
 \begin{itemize}

   \item \textbf{Mean Absolute Error (MAE).}
   It measures the average deviation magnitude:
   \begin{equation}
   \mathrm{MAE} =
   \frac{1}{N}
   \sum_{i=1}^{N}
   \left| S_{\text{pred}}^{(i)} - S_{\text{gt}}^{(i)} \right|
   \end{equation}

   \item \textbf{Adjacent Accuracy (Adj.\ Acc.).}
   It tolerates minor subjective variations by counting predictions within $\pm1$ as acceptable:
   \begin{equation}
   \mathrm{AdjAcc} =
   \frac{1}{N}
   \sum_{i=1}^{N}
   \left(
   \left| S_{\text{pred}}^{(i)} - S_{\text{gt}}^{(i)} \right| \leq 1
   \right)
   \end{equation}

 \end{itemize}
}

In our main results, we report both overall performance and per-dimension breakdown (Aesthetic/Semantic/Spatial/Structural) to highlight diagnostic granularity.

\begin{table*}[!t]
  \centering
  \caption{\textbf{SVGEval task formats.} The same (prompt, rendering, SVG) is evaluated under (a) binary diagnosis questions (Yes/No) and (b) four-aspect scoring on a 1--5 Likert scale.}
  \setlength{\tabcolsep}{3.5pt}
  \renewcommand{\arraystretch}{1.0}

  \begin{subtable}{\linewidth}
    \centering
    \footnotesize

    \begin{tabular}{|@{} M{0.18\linewidth} | L{0.71\linewidth} | M{0.07\linewidth} @{}|}
      \hline
      \multicolumn{1}{|c|}{\textbf{SVG Input}} &
      \multicolumn{1}{c|}{\textbf{Question (Binary Assessment)}} &
      \multicolumn{1}{c|}{\textbf{GT}} \\
      \hline
      \svgthumb{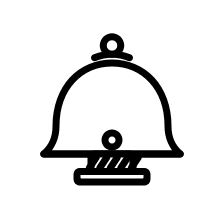} &
      \scriptsize Is the image free of stray marks or unintended scattered strokes outside the main shape, leaving a clean white background? &
      \scriptsize Yes \\
      \hline
      \svgthumb{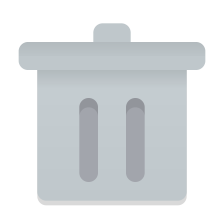} &
      \scriptsize Are the two vertical, rounded openings on the front of the main object at different vertical heights, with one clearly higher than the other? &
      \scriptsize No \\
      \hline
    \end{tabular}

    \vspace{2pt}
    \caption{Task 1: Yes/No Binary Classification}
  \end{subtable}

  \vspace{3pt}

  \begin{subtable}{\linewidth}
    \centering
    \footnotesize

    \begin{tabular}{|@{} M{0.18\linewidth} | L{0.40\linewidth} | M{0.16\linewidth} | M{0.22\linewidth} @{}|}
      \hline
      \multicolumn{1}{|c|}{\textbf{SVG Input}} &
      \multicolumn{1}{c|}{\textbf{Text Prompt}} &
      \multicolumn{1}{c|}{\textbf{\shortstack{Dimension}}} &
      \multicolumn{1}{c|}{\textbf{Score}} \\
      \hline
      \svgthumb{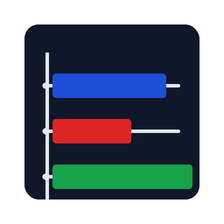} &
      \scriptsize an SVG icon \dots a horizontal bar chart with three bars \dots bold primary colors \dots sharp, clean lines. &
      \textbf{Structural} &
      \textbf{Score: 4} \\
      \hline
    \end{tabular}

    \vspace{2pt}
    \caption{Task 2: Multi-dimensional Scoring}
  \end{subtable}

  \label{Table:benchmark_overview}
\end{table*}

\section{Explainable SVG Quality Scorer}

\myrunin{Goal}
We aim to provide a human-aligned scorer for perceptual-quality evaluation in text-to-SVG generation.
Beyond producing accurate numeric scores, the scorer is required to output \textbf{evidence-grounded rationales} that match expert judgment logic, enabling reliable diagnosis and actionable feedback.
Accordingly, our training target is a mapping from \textbf{(prompt, rendered image, SVG code, dimension)} to \textbf{(dimension-specified score, rationale)} under the four SVGEval dimensions.

\subsection{Reasoning-Distilled Data Synthesis}
\myrunin{Motivation}
Human annotations naturally provide gold scores but rarely include detailed, structured rationales.
To equip the scorer with explainable evaluation capability, we construct high-quality \textbf{image--rationale--score} triples via \textbf{reasoning distillation}.

\myrunin{Score-Conditioned Rationalization}
Despite the availability of high-quality \textbf{human gold scores}, our experiments show that score-only training does not fully exploit the data: performance gains are limited and the resulting model behaves as a black-box regressor with weak diagnostic capability(see \cref{ablation}).
To address this, we additionally generate \textbf{evidence-grounded rationales}, and construct image--rationale--score supervision via reasoning distillation.
Given the prompt, rendered image, SVG code, a target dimension, and its gold score, a teacher model (we apply Qwen3-VL-235B-A22B) is instructed to act as an \emph{analyst}:
\emph{``The score on dimension $d$ is $s$ (assigned by human experts). Identify concrete visual/code evidence that supports this score according to the rubric.''}

\myrunin{Dimension-Constrained Rationales}
To reduce cross-dimension interference, we enforce \textbf{dimension-specific constraints} in the synthesis prompts.
For example, when generating rationales for \textit{Semantic}, the teacher is instructed to ignore visual refinement issues (Aesthetic) and geometry/layout defects (Structural/Spatial), focusing only on concept alignment with the prompt.
We further incorporate the SVGEval hierarchical attribution principle, ensuring that synthesized rationales follow the same diagnostic logic as the benchmark.

\subsection{Model Architecture and Training Recipe}
\myrunin{Base Model and Inputs}
According to the results of our benchmark in \cref{bench_res}, we adopt a strong open-source multimodal model as the backbone (Qwen3-VL 30B A3B).
To leverage the dual nature of SVG and LLM's ability to analyze code, the scorer takes both the \textbf{rendered image} and the \textbf{SVG source code} as inputs, together with the text prompt.
This design allows the model to cross-validate perceptual evidence (e.g., broken strokes, unintended overlaps) with code-level cues (e.g., unclosed paths, anomalous coordinates), which turns out to improve relative capabilities of our scorer (\cref{ablation}).

\myrunin{Output Format}
The scorer outputs \textbf{dimension-specified} results, consisting of a numeric score on a \textbf{1--5 Likert scale} and a short evidence-grounded rationale.
A fixed structured format is used to facilitate automatic parsing and evaluation.

\myrunin{Training}
We perform supervised fine-tuning on the synthesized dataset.
The training objective minimizes the generation loss of both the rationale text and the corresponding score tokens, encouraging the model to produce \textbf{faithful, rubric-aligned explanations} alongside calibrated scores.
At inference time, the scorer produces specified-aspect score with explicit rationale, enabling interpretable assessment of SVG generation quality.

\begin{figure*}[!t]
  \centering
  \begin{subfigure}{0.47\textwidth}
    \centering
    \includegraphics[width=0.97\linewidth]{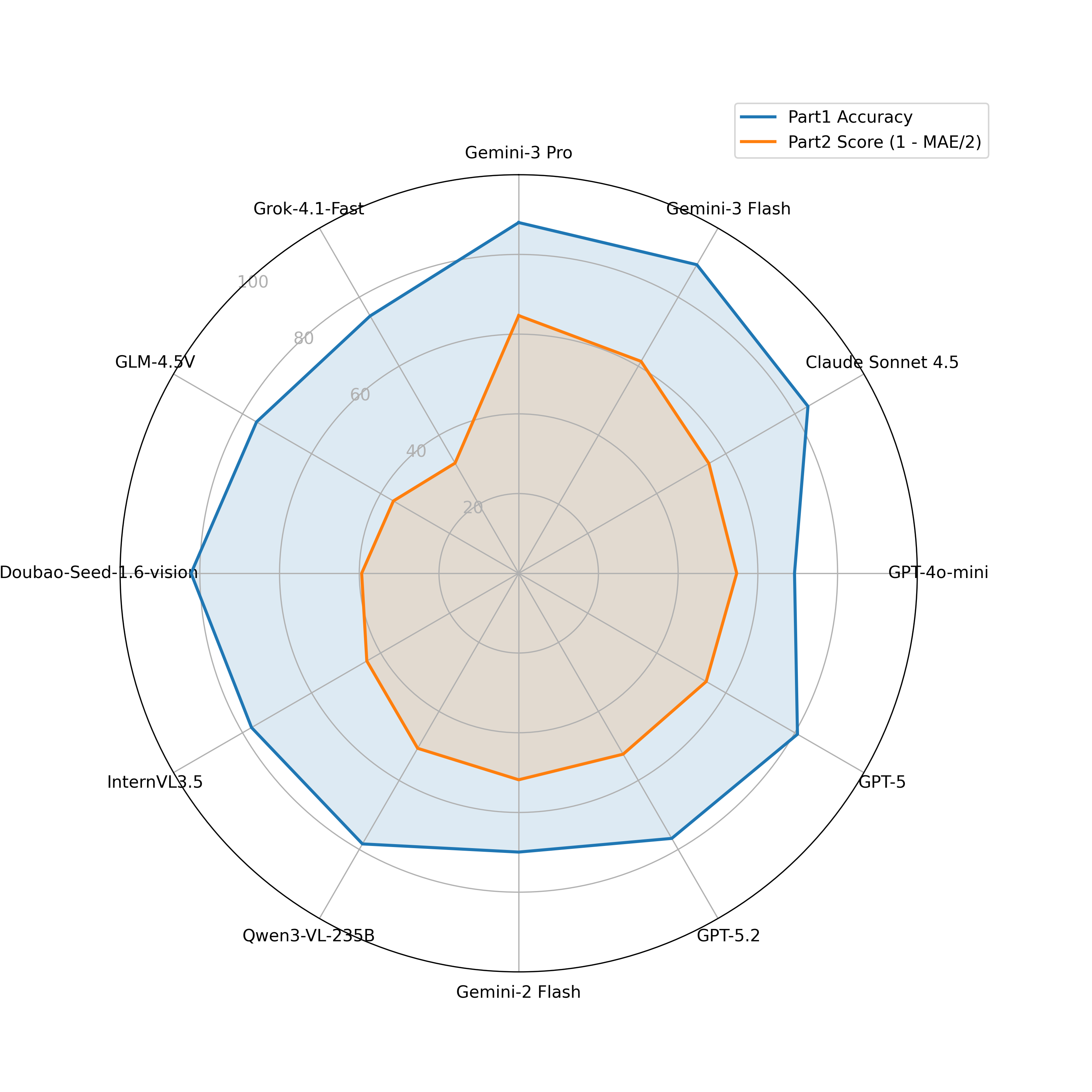}
    \caption{\textbf{Overall performance} on Part I (accuracy) and Part II (visualized from MAE; see text).}
    \label{fig:bench_res}
  \end{subfigure}\hfill
  \begin{subfigure}{0.47\textwidth}
    \centering
    \includegraphics[width=0.97\linewidth]{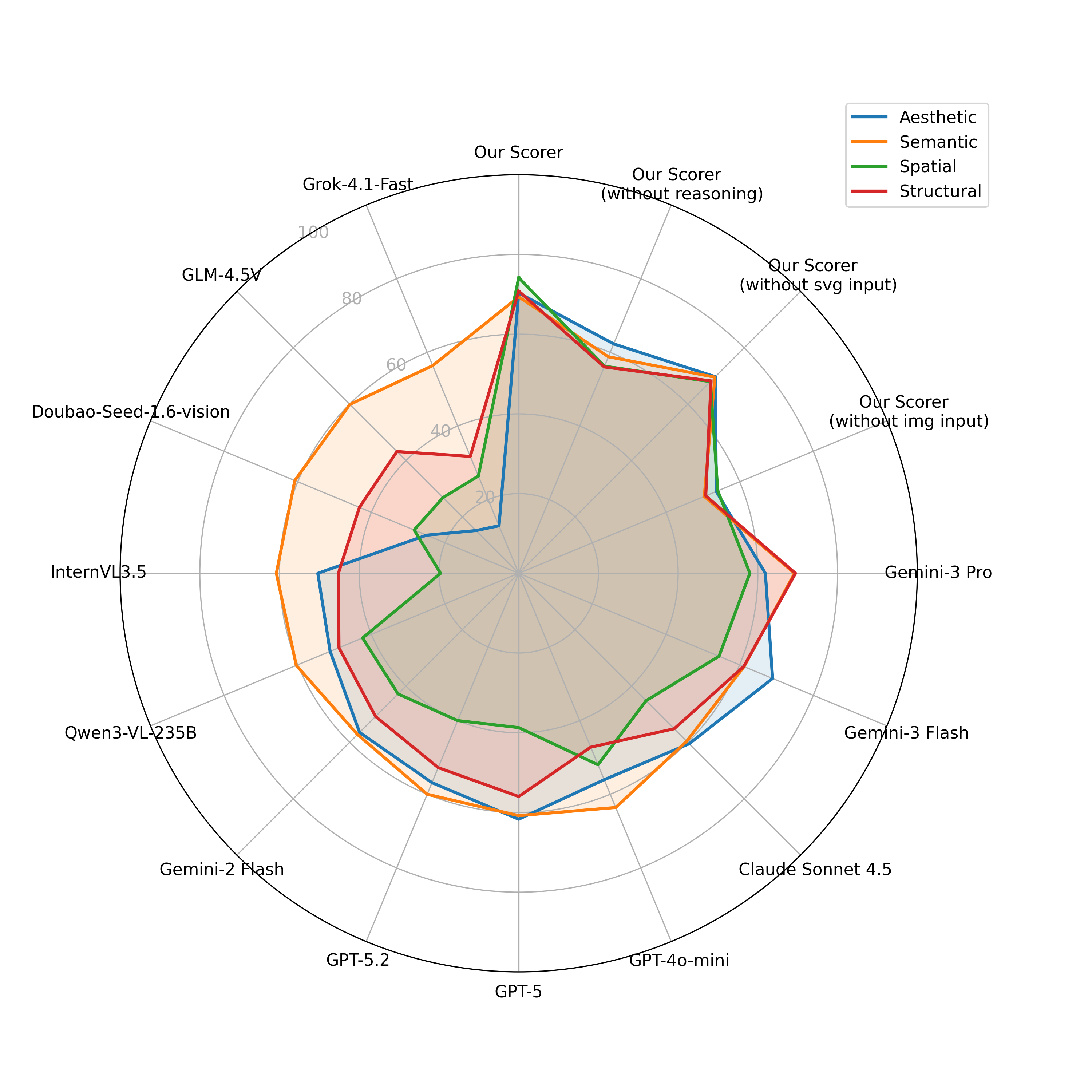}
    \caption{\textbf{Per-dimension performance} on Part II, including SVGEval-Scorer and ablations.}
    \label{fig:scorer_res}
  \end{subfigure}
  \caption{\textbf{Benchmark results on SVGEval.} Models are strong on semantic/aesthetic judgments but weaker on SVG-specific spatial/structural evaluation; SVGEval-Scorer closes the gap, especially on spatial and structural aspects.}
  \label{fig:results}
\end{figure*}

\section{Experiments}

\noindent
We evaluate SVGEval from two perspectives.
First, we benchmark representative multimodal large models to quantify their performance on binary perceptual diagnosis (Part I) and fine-grained multi-aspect scoring (Part II).
Second, we train an explainable SVG quality scorer and analyze the contribution of visual grounding, SVG-code cues, and rationale supervision through controlled ablations.
Together, these experiments aim to (i) diagnose current limitations of MLLMs on SVG-specific structured assessment, and (ii) validate the effectiveness of a dedicated, vision-grounded evaluation model.
Moreover, the benchmarking results also guide our choice of backbone: we select the strongest open-source model under SVGEval as the base model for training the SVG quality scorer.

\subsection{Experimental Setup}
SVGEval evaluates two complementary capabilities: (i) \textbf{Part I} perceptual diagnosis with binary Yes/No judgments, and (ii) \textbf{Part II} multi-aspect quality assessment on a \textbf{1--5 Likert scale}.
For Part I, we report \textbf{Accuracy}.
For Part II, we report \textbf{MAE} and \textbf{Adjacent Accuracy (Adj.\ Acc.)}, where a prediction is counted as correct if it falls within $\pm 1$ of the human gold score.
Unless otherwise specified (ablations), evaluators are provided with the \textbf{text prompt, rendered image, and SVG code}.
Notably, gold labels and rubrics are defined with respect to the \textbf{rendered outcome}.

\begin{table}[!t]
\caption{Overall results on SVGEval. Binary Diagnosis task (Part1) reports binary accuracy. Scoring task (Part2) reports MAE and Adjacent Accuracy on the 1--5 scale.Our SVGEval-Scorer is designed for the multi-aspect scoring task (Scoring); therefore Binary Diagnosis task is not applicable and results are marked with “—”.}
\label{tab:svgeval_overall}
\centering
\footnotesize
\setlength{\tabcolsep}{4pt}
\renewcommand{\arraystretch}{0.95}
\begin{tabularx}{\linewidth}{@{}Xccc@{}}
\toprule
\textbf{Model} &
\makecell{\textbf{Binary Diagnosis}\\\textbf{Acc. (\%) $\uparrow$}} &
\makecell{\textbf{Scoring}\\\textbf{MAE $\downarrow$}} &
\makecell{\textbf{Scoring}\\\textbf{Adj.\ Acc. (\%) $\uparrow$}} \\
\midrule
Gemini-3 Flash & 89.40 & 0.7720 & 62.80 \\
Gemini-3 Pro   & 88.00 & 0.7068 & 66.57 \\
Claude Sonnet 4.5 & 83.80 & 0.8980 & 57.90 \\
Doubao-Seed 1.6 Vision & 82.20 & 1.2120 & 46.20 \\
GPT-5 & 80.76 & 0.9137 & 57.13 \\
Qwen3-VL-235B & 78.40 & 0.9860 & 53.90 \\
InternVL3.5 & 77.40 & 1.1200 & 50.50 \\
GPT-5.2 & 76.80 & 0.9520 & 55.50 \\
GLM-4.5V & 75.95 & 1.2740 & 44.79 \\
Grok-4.1-Fast & 74.55 & 1.3614 & 42.47 \\
Gemini-2.0 Flash & 69.94 & 0.9640 & 54.90 \\
GPT-4o-mini & 69.20 & 0.9060 & 56.80 \\
\midrule
\textbf{SVGEval-Scorer (ours)} & -- & \textbf{0.5772} & \textbf{71.95} \\
\quad w/o reasoning (\texttt{nothink}) & -- & 0.8330 & 60.30 \\
\quad w/o SVG input (\texttt{nosvg}) & -- & 0.6220 & 70.40 \\
\quad w/o image input (\texttt{noimg}) & -- & 0.9557 & 57.75 \\
\bottomrule
\end{tabularx}
\end{table}

\subsection{Benchmark Results}
\label{bench_res}
\myrunin{Overall performance}
\cref{tab:svgeval_overall,fig:bench_res} summarize overall results on SVGEval.
For radar visualization, we convert MAE to a higher-is-better score using a monotonic mapping (e.g., $\mathrm{Score}_{vis}=1-\mathrm{MAE}/2$) and linearly rescale it to $[0,100]$.
This mapping is \textbf{only used for visualization}; all quantitative comparisons and conclusions are based on MAE and Adj.\ Acc.

We observe that modern MLLMs achieve reasonably high accuracy on the binary diagnosis task (Part I; \textbf{69.2--89.4\%}), whereas calibrated multi-aspect scoring remains challenging (Part II; \textbf{MAE 0.71--1.36} and \textbf{Adj.\ Acc. 42.5--66.6\%} for general-purpose models in \cref{tab:svgeval_overall}). 
Notably, performance on Part I and Part II is not perfectly correlated, suggesting that detecting discrete defects does not directly translate to producing well-calibrated ordinal ratings.

Across dimensions, models exhibit substantially larger errors on the two SVG-specific aspects, \textbf{Spatial} and \textbf{Structural}, than on \textbf{Aesthetic} and \textbf{Semantic} (\cref{tab:svgeval_dim_mae}); \textbf{Spatial} is most frequently the hardest dimension, reflecting persistent difficulty in layout/topology reasoning.
Meanwhile, the relatively stronger dimensions (\emph{Semantic} and \emph{Aesthetic}) align with the primary focuses of many raster-centric MLLM benchmarks (e.g., \cite{mme,qbench,qbenchplus}), suggesting that such perceptual skills are comparatively better studied and steadily improved.
In contrast, SVGEval exposes a clear gap on SVG-critical \emph{Spatial} and \emph{Structural} dimensions, indicating that geometry integrity and layout/topology reasoning remain challenging for vector graphics.

\begin{table}[t]
\caption{Per-dimension MAE on Scoring task (Part II) (1--5 Likert).}
\centering
\scriptsize
\setlength{\tabcolsep}{3pt}
\renewcommand{\arraystretch}{0.8}
\begin{tabular}{lccccc}
\toprule
Model & MAE $\downarrow$ & Aesthetic $\downarrow$ & Semantic $\downarrow$ & Spatial $\downarrow$ & Structural $\downarrow$ \\
\midrule
GPT-4o-mini & 0.9060 & 0.8790 & 0.7280 & 0.9600 & 1.0556 \\
GPT-5 & 0.9137 & 0.7661 & 0.7840 & 1.2258 & 0.8800 \\
Gemini-3 Flash & 0.7720 & 0.6210 & 0.7760 & 0.9120 & 0.7778 \\
Qwen3-VL-235B & 0.9860 & 0.9758 & 0.7920 & 1.1520 & 1.0238 \\
Doubao-Seed-1.6-Vision & 1.2120 & 1.5000 & 0.7840 & 1.4320 & 1.1349 \\
Grok-4.1-Fast & 1.3614 & 1.7419 & 0.8720 & 1.4715 & 1.3651 \\
Gemini-2.0 Flash & 0.9640 & 0.8710 & 0.8560 & 1.1440 & 0.9841 \\
InternVL3.5 & 1.1200 & 0.9919 & 0.7840 & 1.6080 & 1.0952 \\
Gemini-3 Pro & 0.7068 & 0.7623 & 0.6160 & 0.8400 & 0.6111 \\
GLM-4.5V & 1.2740 & 1.6967 & 0.8017 & 1.4628 & 1.1360 \\
Claude Sonnet 4.5 & 0.8980 & 0.7903 & 0.8080 & 1.0960 & 0.8968 \\
GPT-5.2 & 0.9520 & 0.8629 & 0.8000 & 1.2000 & 0.9444 \\
\midrule
\textbf{SVGEval-Scorer (ours)} & \textbf{0.5772} & \textbf{0.5935} & \textbf{0.6129} & \textbf{0.5167} & \textbf{0.5840} \\
\quad w/o reasoning (\texttt{nothink}) & 0.8330 & 0.7540 & 0.8240 & 0.8750 & 0.8790 \\
\quad w/o SVG input (\texttt{nosvg}) & 0.6220 & 0.6048 & 0.6080 & 0.6400 & 0.6349 \\
\quad w/o image input (\texttt{noimg}) & 0.9557 & 0.9268 & 0.9920 & 0.9187 & 0.9841 \\
\bottomrule
\end{tabular}
\label{tab:svgeval_dim_mae}
\end{table}

\begin{table*}[!t]
\caption{\textbf{Example outputs of SVGEval-Scorer.} Representative rendered SVGs, evaluated dimensions/scores, and evidence-grounded rationales. Prompts and rationales are truncated for readability; highlighted phrases are model-identified evidence.}
\label{tab:scorer_examples}
\centering
\scriptsize
\setlength{\tabcolsep}{2.6pt}
\renewcommand{\arraystretch}{0.9}

\begin{tabular}{@{}
M{0.13\textwidth}
@{\hspace{6pt}}
M{0.09\textwidth}
@{\hspace{10pt}}
L{0.3\textwidth}
@{\hspace{10pt}}
L{0.43\textwidth}
@{}}
\toprule
\textbf{Rendered SVG} &
\makecell[c]{\textbf{Dim.}\\\textbf{/ Score}} &
\textbf{Prompt (truncated)} &
\textbf{Rationale (truncated)} \\
\midrule

\includegraphics[width=0.72\linewidth]{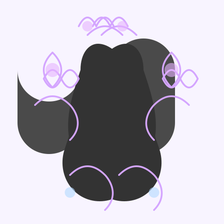} &
\makecell{\tiny\textbf{Aesthetic}\\[-1pt]\textbf{1/5}} &
a woman’s silhouette with flowing hair, \dots\ soft pastel colors, \dots\ conveying feminine elegance and tranquility. &
a silhouette that lacks clarity and definition \dots\ overly simplified and becomes indistinct. \dots\
The purple/blue hues appear randomly placed and fail to form a harmonious composition. \dots\
\textbf{visually noisy} \dots\ result in a very poor aesthetic quality. \\

\addlinespace[0pt]

\includegraphics[width=0.72\linewidth]{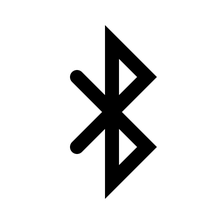} &
\makecell{\tiny\textbf{Spatial}\\[-1pt]\textbf{5/5}}  &
a black Bluetooth symbol: a stylized ``B'' formed by \dots\ overlapping diagonal lines, thick solid black strokes \dots &
Bluetooth symbol \dots\ \textbf{excellent spatial coherence}. \dots\ two diagonal lines forming the ``B'' \dots\ match what is expected for a well-designed Bluetooth icon \dots \\

\addlinespace[0pt]

\includegraphics[width=0.72\linewidth]{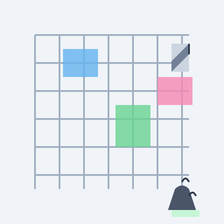} &
\makecell{\tiny\textbf{Semantic}\\[-1pt]\textbf{3/5}} &
open planner with weekly grid \dots\ colored appointment blocks, a pencil icon, and a coffee cup in the corner. &
grid and blocks are present\dots\ a pencil icon appears as required. \dots\
a vague cup-like shape is visible \dots\ the coffee cup is \textbf{not clearly depicted} \dots\
slightly messy layout \dots\ justifying a mid-level score. \\

\addlinespace[0pt]

\includegraphics[width=0.72\linewidth]{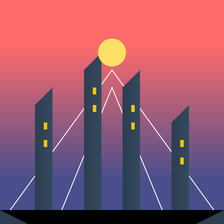} &
\makecell{\tiny\textbf{Structural}\\[-1pt]\textbf{4/5}} &
cityscape\dots\ receding skyscrapers, vanishing-point lines \dots\ dynamic shadows\dots\ sunset gradient sky. &
clean lines and well-defined structures\dots\
perspective lines converge neatly toward a central vanishing point. \dots\
\textbf{slightly coarse shadows} reduce structural precision, but overall execution remains high-quality. \\

\addlinespace[0pt]

\includegraphics[width=0.72\linewidth]{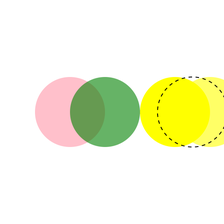} &
\makecell{\tiny\textbf{Structural}\\[-1pt]\textbf{2/5}} &
two sets of overlapping circles connected by an arrow \dots\ &
\dots\ closed and non-intersecting path \dots\
the design is \textbf{overly simplistic} \dots\  too basic to warrant a higher score. \\

\addlinespace[0pt]

\includegraphics[width=0.72\linewidth]{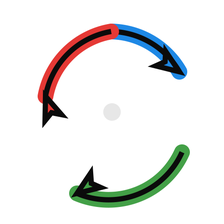} &
\makecell{\tiny\textbf{Semantic}\\[-1pt]\textbf{4/5}} &
 three intertwined arrows in red, blue, and green, forming a circular motion to symbolize recycling \dots\ &
The arrows are intertwined \dots\
symbolizes recycling \dots\ which is consistent with the description \dots
However, the arrows \textbf{do not perfectly form a continuous circular motion} \dots\ \\

\bottomrule
\end{tabular}
\vspace{-4pt}
\end{table*}

\subsection{Explainable SVG Quality Scorer}
\label{ablation}
\myrunin{Main results}
We train \textbf{SVGEval-Scorer} to output dimension-wise scores together with evidence-grounded rationales.
As shown in \cref{tab:svgeval_overall}, SVGEval-Scorer substantially improves Part II evaluation, achieving lower MAE and higher Adj.\ Acc than directly using general-purpose models as evaluators.
The improvement is consistent across all dimensions, with particularly strong gains on \textbf{Spatial} and \textbf{Structural} (\cref{tab:svgeval_dim_mae,fig:scorer_res}). We attribute the gain to (i) explicit visual grounding via renderings, (ii) complementary structural cues from SVG code, and (iii) rationale supervision; we validate each factor with targeted ablations below.

\myrunin{Ablation study}
We conduct three ablations to assess the role of modalities and rationale supervision:

{\setlength{\topsep}{2pt}
 \setlength{\itemsep}{2pt}
 \setlength{\parsep}{0pt}
 \begin{itemize}
   \item \textbf{w/o reasoning (\texttt{nothink}).}
   Removing the rationale generation objective causes a clear degradation (\cref{tab:svgeval_overall,tab:svgeval_dim_mae}).
   Beyond the metric drop, the scorer becomes substantially less \textbf{auditable}: correct scores may be incidental, while failures are difficult to diagnose without evidence-grounded attributions, effectively reducing the evaluator to a black-box regressor.

   \item \textbf{w/o SVG input (\texttt{nosvg}).}
   Removing SVG code only mildly affects overall performance, but leads to more noticeable drops on \textbf{Spatial} and \textbf{Structural} (\cref{tab:svgeval_dim_mae}).
   This suggests that the \textbf{rendered image dominates} perception-based assessment, while SVG code provides \textbf{complementary structural cues} (e.g., path closure, anomalous coordinates, layer ordering) that are particularly helpful for SVG-critical dimensions.

   \item \textbf{w/o image input (\texttt{noimg}).}
   When only SVG code is available, performance degrades sharply (MAE $0.9557$, Adj.\ Acc.\ $57.75\%$ in \cref{tab:svgeval_overall}).
   We observe that the model often collapses to \textbf{format imitation} with unstable score calibration (frequent large-score deviations) and tends to produce less reliable evidence statements without direct visual grounding, highlighting the necessity of renderings for perceptual SVG quality assessment.
 \end{itemize}
}

\section{Conclusion}
We presented \textbf{SVGEval}, a vision-grounded framework for perceptual-quality benchmarking in text-to-SVG generation. 
SVGEval defines a human-aligned four-aspect rubric (\emph{Aesthetic, Semantic, Spatial, Structural}) and constructs a two-part benchmark with mixed task formats (binary diagnosis and 1--5 Likert scoring), where evaluators jointly access the prompt, rendered image, and SVG code while gold labels are defined over the rendered outcome. 
Systematic evaluations across representative MLLMs reveal a consistent gap: models are relatively stronger at \emph{Semantic} and \emph{Aesthetic} judgments but struggle on SVG-critical \emph{Spatial} and \emph{Structural} assessment.
Building on SVGEval, we trained an \textbf{explainable SVG quality scorer} with explicit visual grounding and rationale supervision, achieving substantially improved calibration and robustness, especially on spatial/layout and geometric integrity.
We hope SVGEval can serve as a reliable testbed and a practical training signal for developing structure-aware SVG generation and evaluation systems.

\section{Limitations}
First, although SVGEval adopts multi-round human annotation with expert adjudication, fine-grained perceptual judgments on a 1--5 Likert scale are inherently subjective. 
Minor disagreements may remain for borderline cases, especially for aesthetics-related scores, and it may be possible to model it as expected perceptual variance to reduce subjective noise.

Second, SVGEval defines gold labels over the rendered outcome under a fixed rasterization pipeline for consistency. 
In occasional cases, subtle rasterization artifacts may influence model judgments when evaluating structural precision. 
This phenomenon indicates that more explicit image–SVG cross-referenced evaluation strategies could further improve robustness, by jointly considering perceptual evidence from the rendering and structural cues from the underlying vector code.

\paragraph{Acknowledgements.} 
\vspace{1.2\baselineskip}
This work is financially supported by the Open Research Fund from Guangdong Laboratory of Artificial Intelligence and Digital Economy (SZ), under Grant NO.GML-KF-26-07, and it is supported by the Science
and Technology Commission of Shanghai Municipality under research grant No. 25ZR1401187.

\clearpage

%
%
\bibliographystyle{splncs04}
\bibliography{main}
\end{document}